\documentclass[twoside,11pt]{article}
\usepackage{jair, theapa, rawfonts}
\usepackage{amsmath, amssymb, amsfonts, amsthm}
\usepackage{array}
\usepackage{multirow}
\usepackage{algorithmic}
\usepackage{graphicx}
\usepackage{textcomp}
\usepackage{xcolor}
\usepackage{url}

\newtheorem{theorem}{Theorem}
\newtheorem{lemma}[theorem]{Lemma}
\newtheorem{definition}{Definition}

\jairheading{82}{2025}{209-240}{06/2024}{01/2025}
\ShortHeadings{Confidence-based Estimators for Predictive Performance in Model Monitoring}
{Kivimäki, Bia{\l}ek, Nurminen \& Kuberski}
\firstpageno{209}

\begin{document}

\title{Confidence-based Estimators for 

Predictive Performance in Model Monitoring}

\author{\name Juhani Kivimäki \email juhani.kivimaki@helsinki.fi \\
       \name Jukka K. Nurminen \email jukka.k.nurminen@helsinki.fi \\
       \addr University of Helsinki
       \AND
       \name Jakub Bia{\l}ek \email jakub@nannyml.com \\
       \name Wojtek Kuberski \email wojtek@nannyml.com \\
       \addr NannyML
}

\maketitle

\begin{abstract}
After a machine learning model has been deployed into production, its predictive performance needs to be monitored. Ideally, such monitoring can be carried out by comparing the model's predictions against ground truth labels. For this to be possible, the ground truth labels must be available relatively soon after inference. However, there are many use cases where ground truth labels are available only after a significant delay, or in the worst case, not at all. In such cases, directly monitoring the model's predictive performance is impossible. 

Recently, novel methods for estimating the predictive performance of a model when ground truth is unavailable have been developed. Many of these methods leverage model confidence or other uncertainty estimates and are experimentally compared against a naive baseline method, namely Average Confidence (AC), which estimates model accuracy as the average of confidence scores for a given set of predictions. However, until now the theoretical properties of the AC method have not been properly explored. In this paper, we bridge this gap by reviewing the AC method and show that under certain general assumptions, it is an unbiased and consistent estimator of model accuracy. We also augment the AC method by deriving valid confidence intervals for the estimates it produces. These contributions elevate AC from an ad-hoc estimator to a principled one, encouraging its use in practice.

We complement our theoretical results with empirical experiments, comparing AC against more complex estimators in a monitoring setting under covariate shift. We conduct our experiments using synthetic datasets, which allow for full control over the nature of the shift. Our experiments with binary classifiers show that the AC method is able to beat other estimators in many cases. However, the comparative quality of the different estimators is found to be heavily case-dependent. 
\end{abstract}

\section{Introduction}
\label{Introduction}

The last decade has seen a considerable improvement in the predictive capability of machine learning (ML) models. However, as techniques in designing and implementing these models have matured, their use in real-life scenarios is still in its infancy. There are many reasons for the slow adoption of ML models in industrial practices. The real-life use cases require the models to be integrated with other services and data pipelines, forming a complex \textit{Machine Learning system}~\shortcite{schroder:2022}. Furthermore, questions of reliability need to be addressed.

It is widely accepted that ML models are rarely perfect in their predictions, which always contain some amount of uncertainty. However, in many cases, especially in safety-critical ones, one would like to be able to quantify these uncertainties reliably. Often it is not sufficient to know what the average predictive performance for a held-out test set was in the design phase of the model. One also needs to monitor the uncertainties and model performance after deployment. This is ideally done on a batch or even instance level. 

Monitoring can be directed towards many different stages in an ML system. Roughly speaking, one can monitor input data~\shortcite{rabanser:2019}, the inner workings of a model~\shortcite{myllyaho:2022}, or model outputs \shortcite{hendrycks:2017,corbiere:2019}. The monitoring can be targeted to capture anomalies, drifts in the data, or performance deterioration~\shortcite{klaise:2020}. One could even extend the scope of monitoring to capture aspects such as server throughput or even business impact. In this work, we focus only on methods that monitor the predictive performance of models based on their outputs.

In some cases, an ML practitioner can compare the predictions of a model with the actual ground truth (GT) values with some acceptable latency. For example, if a model is used to predict tomorrow's stock prices, one can assess the quality of those predictions with a lag of one day. In some cases, the GT values for a portion of predictions can be queried from an expert~\shortcite{ginart:2022}. In such cases, monitoring predictive performance is relatively straightforward~\shortcite{klaise:2020} and one can choose from a wide range of algorithms~\shortcite{bayram:2022}.

Unfortunately, there are plenty of cases where GT becomes available only after a substantial lag or not at all. In such cases, one can not directly monitor the predictive performance of an ML system with acceptable latency. Historically in these situations, only shifts in the data have been monitored with the hope that they are indicative of shifts in model performance. However, this is not always the case, and making such assumptions might result in unnecessary re-trainings inducing unnecessary costs~\shortcite{baier:2021}. This dilemma becomes even worse with high-dimensional data, although using techniques such as dimensionality reduction~\shortcite{rabanser:2019} can alleviate this to some extent. Clearly, tools for distinguishing meaningful shifts from redundant ones are needed.

Only quite recently, there have been several propositions in how to solve the \textit{unsupervised accuracy estimation} problem, where instead of monitoring shifts in the data, one tries to estimate the predictive performance directly~\shortcite{baek:2022,chen:2021,deng:2023,garg:2022,guillory:2021,jiang:2022,lu:2023} in the absence of GT. Many of the papers proposing these new estimation methods use the \textit{Average Confidence} (AC)~\shortcite{hendrycks:2017} of a model as a simple baseline for estimating the predictive accuracy and experimentally show their method to beat this baseline under different scenarios. However, none of these papers explain in detail, why they choose to use AC as a comparative baseline, nor present the theoretical properties of this baseline. 

In this paper, we present a justification for using AC as a baseline method for estimating the predictive performance of an ML system in detail. We prove theoretically that under certain assumptions, this method is an unbiased and consistent estimator of the model's predictive performance. Furthermore, we complement and contrast some of the results presented in earlier papers \shortcite{guillory:2021,garg:2022} by performing empirical tests under different scenarios. 

The key contributions of this paper are:
\begin{enumerate}
    \item We analyze the theoretical properties of AC, a baseline method for estimating the accuracy of a deployed machine learning model in the absence of ground truth labels during inference, and provide theoretical guarantees for the quality of its estimates under the so-called calibration assumption (in Section~\ref{method}).
    \item We show that by leveraging properties of the Poisson binomial distribution, one can derive valid confidence intervals for the estimates of AC, and demonstrate the quality of the derived confidence intervals experimentally (in Section~\ref{method}). 
    \item We demonstrate that when the calibration assumption is approximately met, AC can deliver better results than some more complex methods described in papers, where the authors have made contrasting claims. (in Section~\ref{experiments}).
    \item We demonstrate that in many cases the estimation error of AC and other confidence-based estimators is highly (linearly) correlated with the calibration error (in Section~\ref{experiments}).
\end{enumerate}

\section{Background}\label{background}

In this section, we present the necessary background theory for confidence-based model monitoring. We begin with definitions for different distributional shifts, then shortly describe a standard statistical approach to model monitoring when GT is available, and finally discuss confidence scores and model calibration.

\subsection{Unsupervised Accuracy Estimation under Dataset Shift}

Let us assume that we have trained a supervised ML model $f$ using data from a \textit{source distribution} $p_s(\boldsymbol{x},y)$ available to us during development. Once the model is deployed, it will make predictions using data from a \textit{target distribution} $p_t(\boldsymbol{x},y)$. Typically $p_s(\boldsymbol{x},y) \neq p_t(\boldsymbol{x},y)$. Following the terminology of~\shortciteA{moreno-torres:2012}, we label this inequality as \textit{dataset shift}. 

During development, we have evaluated the performance of our model $f$ with unseen data from $p_s(\boldsymbol{x},y)$ and would like to evaluate the performance with data from $p_t(\boldsymbol{x},y)$. However, since we have access only to $p_t(\boldsymbol{x})$, evaluating the performance on $p_t(\boldsymbol{x},y)$ directly is not possible. Thus at best, we can only \textit{estimate} the predictive performance without GT labels.

For an estimator to work, assumptions about the nature of the shift have to be made~\shortcite{garg:2022}. Since for any distribution $p(\boldsymbol{x},y) = p(y|\boldsymbol{x})p(\boldsymbol{x})$, the dataset shift can be caused by \textit{concept shift}, where $p_s(y|\boldsymbol{x}) \neq p_t(y|\boldsymbol{x})$. If this is not the case, it must be caused by \textit{covariate shift}, where $p_s(\boldsymbol{x}) \neq p_t(\boldsymbol{x})$. These causes are taken as mutually exclusive. If both occur, the generic term of dataset shift is used. Alternatively, one can also decompose $p(\boldsymbol{x},y) = p(\boldsymbol{x}|y)p(y)$. In this case, a situation where $p_s(\boldsymbol{x}|y) = p_t(\boldsymbol{x}|y)$ and $p_s(y) \neq p_t(y)$ is called \textit{label shift}. In this paper, we focus on cases of covariate shift. 


The most important takeaway here is that if the model being monitored is fixed and deterministic after training (as is usually the case), persisting changes in model performance in production (assuming the ML system is working properly) can only be attributed to changes in the underlying data distribution. Next, we discuss how to identify if changes in model performance are indeed persisting or just random fluctuations due to sampling effects.

\subsection{Traditional Process Monitoring}
\textit{Statistical Process Control} (SPC) was pioneered by Shewhart at Bell Labs in the 1920s for systematic monitoring and quality control of industrial production processes. This approach has since been applied to all kinds of processes outside manufacturing with great success. In this wider context, a process can be defined as "everything required to turn an input into an output for a customer"~\shortcite{stapenhurst:2005}. As such, the predictions of an ML model fit this pattern as well.

In SPC it is assumed that there is always some variation present in the process being monitored. Furthermore, it is assumed that the variation can be divided into two types, namely \textit{common cause variation} and \textit{special cause variation}. Ideally, a process is only affected by common cause variation. If this is the case, the process is said to be under \textit{statistical control}, meaning that the variance observed in the process conforms to expectations. 

\textit{Control charts} are central tools used to monitor processes in SPC. A control chart is essentially a run chart of some statistic calculated from consecutive samples. It tracks whether the variation of the statistic from the mean stays within preset control limits. There are various established sets of decision rules~\shortcite<such as the Western Electric rules, see e.g.>{montgomery:2009} for recognizing when the data shown on a control chart implies that the process has likely strayed outside statistical control. If this is the case, an alarm is raised for the user. In an ML monitoring setting, widely-used algorithms such as Drift Detection Method (DDM) and its numerous variants~\shortcite{bayram:2022} are implementations of SPC. However, these methods can only be applied with access to GT.

Though there are many variants of control charts for monitoring different processes, in this work we are only interested in two variants, namely $np$-charts and $p$-charts, which are used to monitor the number and fraction of non-conforming units in a sample respectively. Both of these divide each unit in a sample into binary categories of either \textit{conforming} (pass) or \textit{non-conforming} (fail). Naturally, the user needs to establish criteria for treating each unit accordingly. These charts are used in monitoring by gathering samples from the process and tracking the number or fraction of non-conforming units in each sample. The use of these charts requires three assumptions to be made:
\begin{enumerate}
    \item The probability of nonconformity $p$ is the same for each unit.
    \item Each unit is independent of its predecessors or successors.
    \item The inspection procedure is the same for each sample and is carried out consistently from sample to sample.
\end{enumerate}
In essence, under these assumptions, the nonconformity of each unit is considered to be an i.i.d. random variable following a Bernoulli distribution with parameter $p$. Accordingly, the number of non-conforming units in a sample of $n$ units follows the binomial distribution $B(n, p)$.

Since the sample mean is an unbiased and consistent estimator of the population mean of the process, which is the quantity we ultimately care about, all common cause variation can be attributed to sampling effects. According to the Central Limit Theorem, the sampling distribution is approximately normal, making the control limits interpretable. A typical control limit is to raise an alarm if the number or fraction of non-conforming units is three standard deviations above the mean, corresponding to $99.7\%$ probability that an observed sample mean outside the control limits is an indicator of special cause variation. The statistics of the sampling distribution are determined with some in-control reference data. 

\subsection{Confidence Calibration}
In addition to predicting a certain class, most classifier models also output (or can be modified to output) a vector of soft scores with one value for each class. Usually, the values of the vectors are normalized on the interval $[0,1]$, summing to one, which means they form a categorical probability distribution over the label space. The maximum value of the normalized vector can be interpreted as a \textit{confidence score}~\shortcite{guo:2017} for the prediction, where the value $1$ signals maximal confidence. 

In addition to classifiers, confidence scores can be derived for many other model types. Recently, one increasingly popular approach with non-classifier models has been to train an auxiliary model to assign a confidence score for the predictions of the base model~\shortcite{chen:2019,corbiere:2019,devries:2018,kivimaki:2023,shao:2020}. Using such techniques, one can attach a confidence score to practically any type of model as long as criteria for treating each prediction of the base model as either conforming or non-conforming exists. 

In the context of classifier models, it is tempting to interpret a confidence score as the probability that a given instance was predicted correctly. However, it is rather common that the predicted probabilities do not align with empirical probabilities. That is, if one monitors all the predictions that were assigned a certain confidence score, the fraction of correct predictions among those predictions would deviate from the confidence score significantly~\shortcite{guo:2017,zadrozny:2001}. If the confidence scores produced by a model align with empirical probabilities, they are said to be \textit{calibrated}. We will next provide a more formal definition of calibration.

Let us denote the feature space of our model by $\mathcal{X}$, the output space by $\mathcal{Y}$, and the interval of confidence scores [0,1] by $\mathcal{S}$. In this work, we are interested in estimating the predictive performance of a given supervised model $\textbf{f}: \mathcal{X} \rightarrow \mathcal{Y}\times \mathcal{S}$, with $\textbf{f}=(f_y,f_s)$, where $f_y:\mathcal{X} \rightarrow \mathcal{Y}$ and $f_s:\mathcal{X} \rightarrow \mathcal{S}$. We assume that there exist some fixed criteria ${g:\mathcal{Y}\times\mathcal{Y} \rightarrow \{0, 1\}}$ for deciding if a prediction $\hat{y}=f_y(\boldsymbol{x})$ should be treated as non-conforming or conforming. We let the random variables $X$ and $Y$ denote respectively the features and label drawn from a joint data distribution $p(\boldsymbol{x}, y)$, where $\boldsymbol{x}\in \mathcal{X}$ and $y \in \mathcal{Y}$. We further define derived random variables $\hat{Y}=f_y(X)$, $S=f_s(X)$, and $C=g(Y,\hat{Y})$.

If $f_y$ is a classifier model, a natural choice would be to treat a prediction as conforming if and only if it is correct. Other types of models might require less straightforward ways to assess the (binary) quality of their predictions. For example, regressor models might include tolerance in the form of some acceptable margin of error. This very general formulation enables us to reduce a wide variety of supervised learning settings as binary classification. In this binary classification setting, we can now define what it means for the confidence scores of a model (or the model itself) to be \textit{perfectly calibrated}.
\begin{definition}\label{definition}
    Model $\boldsymbol{\mathrm{f}}: \mathcal{X} \rightarrow \mathcal{Y}\times \mathcal{S}$ is perfectly calibrated within distribution $p(\textbf{x},y)$ iff.
    \begin{equation*}
        P_{p(\boldsymbol{x},y)}(C=1 \mid S=s) = s \quad \forall s \in \mathcal{S}.
    \end{equation*}
\end{definition}
For the sake of brevity, in the rest of this paper, we will refer to the conforming predictions as \textit{correct} and to the non-conforming as \textit{incorrect}. In practice, no model can achieve perfect calibration, so for the concept of calibration to be meaningful, we must be able to quantify calibration error. How this should be done is an active discussion within the scientific community ~\shortcite<see e.g.,>{guo:2017,nixon:2019,posocco:2021,vaicenavicius:2019}. One of the currently most used metrics is \textit{Expected Calibration Error} (ECE), which requires the predictions to be binned into $M$ bins according to their confidence scores. The hyperparameter $M$ is set according to user preferences and affects the estimated calibration error. ECE is defined as
\begin{equation}
    ECE = \sum_{m=1}^M\frac{|B_m|}{n}\left|A_m-C_m\right|,
\end{equation}
where $n$ is the total number of predictions, $B_m$ the set of predictions in bin $m$, and $A_m$ and $C_m$ the average accuracy and confidence of predictions in bin $m$ respectively. In the standard version, the bins are assumed to be equiwidth. In this work, however, we choose to use equisize binning, where bin boundaries are adapted so that each bin holds the same number of predictions. This adaptive binning scheme gives rise to \textit{Adaptive Expected Calibration Error} (ACE), which is shown to be more robust with respect to the number of bins hyperparameter~\shortcite{nixon:2019}. We use $M=20$ in our experiments. Other than the choice of binning, ACE is calculated exactly the same way as ECE.

If the confidence scores are not calibrated, a post hoc \textit{calibration mapping} can be applied to reduce the calibration error and make the confidence scores match empirical probabilities more accurately~\shortcite<e.g.,>{alexandari:2020,guo:2017,kull:2017,kull:2019,kumar:2019,Naeini:2015,platt:1999,zadrozny:2001,zadrozny:2002}. There are also ways in training models inducing an implicit calibration effect on the confidence scores~\shortcite<e.g.,>{kumar:2018,mukhoti:2020,muller:2019,ovadia:2019,seo:2019,zhang:2020}, although most of these methods were not originally designed with calibration in mind.

\section{Related Work}\label{related}

In recent years, there have been many propositions on how to perform unsupervised accuracy estimation. Some of these methods leverage model confidence expressed as the maximum of an output vector from the final softmax layer of a neural network~\shortcite{guillory:2021,garg:2022,deng:2023}. Alternatively, the prediction entropy has been used as a measure of uncertainty. Others train several models and try to assess the predictive accuracy by tracking discrepancies in the predictions of those models~\shortcite{baek:2022,chen:2021,jiang:2022}. One approach is to use importance weighting (IW)~\shortcite{lu:2022} to account for the differences between the source and target distributions~\shortcite{chen2021b}. The methods presented by~\shortciteA{guillory:2021} and~\shortciteA{garg:2022} are most similar to AC, so we will explain them in more detail below and focus on them in our experiments. 

In the Difference of Confidences (DoC) method~\shortcite{guillory:2021}, one first measures the difference of the average confidences of a model on a training set and on multiple shifted datasets, which are created from the training dataset by adding different kinds of synthetic noise. Then, an auxiliary regressor model is trained to estimate the base model's accuracy on the shifted datasets using the difference in confidence scores between the training set and the shifted set as the only feature. The authors show experimentally that this approach leads to better generalization on both synthetic and natural shifts when compared to using common distributional distances such as Frechet distance~\shortcite{frechet:1957} and Maximum Mean Discrepancy~\shortcite{muandet:2017} as features~\shortcite{guillory:2021}. They also experiment with using the difference of average entropies (DoE) in place of average confidences arriving at similar results. Furthermore, and most interestingly for our current pursuit, they compare DoC against three simple regression-free baselines: Average Confidence (AC), Average Confidence after calibration with Temperature Scaling (AC TempScaling)~\shortcite{guo:2017}, and subtracting the Difference of Confidences feature value from training accuracy (DoC-Feat). They show that AC is able to beat all of the regression models trained using distributional distances but loses the comparison against DoC, DoE, and Doc-Feat. Most interestingly, they claim that AC performs better than AC TempScaling, implying that calibration might be harmful. 

One shortcoming of the DoC approach is that it requires the creation of several shifted datasets. This might be infeasible in many cases. Furthermore, no guarantees can be given that the regression function learned with the synthetic shifted datasets would work in all cases of real-world shifts. In fact, ~\shortciteA{garg:2022} experimentally show the existence of cases where the DoC approach would fail. Instead of using the difference of confidence, they propose to use Average Thresholded Confidence (ATC), where one first finds a confidence threshold such that the fraction of training samples below that threshold equals the prediction error rate on the training set. Then, for a shifted dataset the estimated accuracy is the fraction of samples above the threshold. This method is shown to beat AC, DoC-Feat, IW and the GDE method described by~\shortciteA{jiang:2022}. No direct comparison between ATC and DoC is made, with the rationale that ATC is regression-free and DoC is not. Contrary to~\shortciteA{guillory:2021}, the authors found that calibration enhances the effectiveness of their method. In fact, for all methods they tested against (including AC and DoC-Feat), calibration improved the estimates in almost all of the test cases. 

On a more general note, all of the methods mentioned in the first paragraph of this section are tested on the dataset level. That is, a model is first trained on one dataset and its accuracy is evaluated on a shifted dataset, where the shifted dataset consists of thousands or even tens of thousands of samples. For the estimates to be useful in most practical monitoring settings, they should be relatively accurate for much smaller sets of samples, in the order of hundreds of samples. None of the methods are tested on such smaller sets of samples. To make matters worse, the variance of estimation error is not explored for any of the methods. Thus, even though the methods might give good estimates on a dataset level, the estimates on smaller batches of samples might express large variance, which could seriously hamper their usability in estimating accuracy in a monitoring setting. Furthermore, no theoretical guarantees for their methods are given by~\shortciteA{baek:2022,deng:2023,guillory:2021}, and even where such guarantees are given, they depend on some highly specific assumptions. This raises the question of how well would the experimental results described in these papers hold in other settings. 

The methods are mainly tested in a multiclass image classification setting~\shortcite<although>[also include some tests with textual data]{garg:2022,chen:2021,lu:2023} with neural networks, and the confidence scores are taken to be the maximum of the softmax output. This typically results in uncalibrated confidence scores~\shortcite{guo:2017} and although all of the methods utilizing these confidence scores also report results for calibrated confidence scores, the calibration procedure is equated with temperature scaling~\shortcite{guo:2017} and the calibration errors for neither the original nor shifted datasets are reported. This makes using AC as a comparative baseline somewhat problematic since the performance of AC might be seriously hampered under miscalibration (as we shall soon see). 

Although it is taken as common knowledge that calibration error tends to increase under covariate shift~\shortcite{ovadia:2019}, this is only shown to hold for softmax classifiers calibrated with temperature scaling~\shortcite{guo:2017}. In fact, methods to ensure low calibration error under covariate shift are being developed~\shortcite{białek2024}. Interestingly, AC baseline is shown by~\shortcite{guillory:2021} to be comparable or better to DoC under training schemes, which contain data augmentations, such as AugMix~\shortcite{hendrycks:2020} and DeepAugment~\shortcite{hendrycks:2021}. Such augmentation schemes are known to produce implicit calibration for the trained models~\shortcite{hendrycks:2020}. This raises the question of whether the AC baseline would have been more performant under some other calibration scheme. 
 
\section{Average Confidence}\label{method}

In this section, we present some theoretical results for the AC method, which was originally proposed as a baseline method for detecting misclassified and out-of-distribution instances in neural networks~\shortcite{hendrycks:2017}. When used in unsupervised accuracy estimation, the predictive accuracy is estimated to be the average of confidence scores, hence the name.

Our analysis hinges on the assumption that the monitored model produces calibrated confidence scores in addition to its predictions. If this \textit{calibration assumption} is met, the correctness of each prediction can be regarded as a Bernoulli trial where the confidence score $S$ for the prediction acts as the parameter. Assuming that these trials are independent of each other, the number of correct predictions in $n$ predictions is the sum of the outcomes of the $n$ trials, which follows a Poisson binomial distribution, with the confidence scores $S_i$ as parameters. The probability mass function of a random variable $K$ following this distribution is
\begin{equation}
    P(K=k)=\sum_{A\in F_k}\prod_{i\in A}S_i\prod_{j\notin A}(1-S_j),
\end{equation}
where $F_k$ is the set of all subsets of $\{1,2,...,n\}$ with $k$ members. 

Since $|F_k| = \binom{n}{k}$, calculating the required probabilities using the definition directly is infeasible even for relatively small values of $n$. Fortunately, there are several ways to avoid this combinatorial explosion. In this work, we leverage the algorithm presented by~\shortciteA{hong:2013}, which is based on the Fast Fourier Transform and implemented in the \texttt{poibin}\footnote{https://github.com/tsakim/poibin} python library. It allows us to derive the Poisson binomial distributions exactly and almost instantly.

Strictly speaking, our calibration assumption requires the model being monitored to be perfectly calibrated, which (as already stated) is not possible in practice. We will omit this detail in the following theoretical considerations and assume that these results are approximately applicable when the calibration error is small enough. We will return to this issue in Section~\ref{experiments} for more insight on practical applicability.

\subsection{Estimating Predictive Accuracy}\label{accuracy}

The key benefit of leveraging the Poisson binomial distribution to monitor a deployed ML model is that it allows us to waive the binomial assumption used in $p$-charts and $np$-charts, where the non-conformity of each unit is considered equally probable. Instead, we can assign these probabilities for each unit individually and still perform rigorous statistical inference in this more general and complex setting.

In what follows, we assume for simplicity that the monitor is run in a batch fashion where predictions are gathered into non-overlapping fixed-size windows. The window size $n$ can be set by the user according to their needs. Furthermore, we assume that the model being monitored produces calibrated confidence scores and that we do not have access to GT. After each batch of $n$ predictions, the observed confidence scores are used to form a Poisson binomial distribution for the number of correct predictions within that batch. Next, we will show how one can produce unbiased and consistent estimates for predictive accuracy under these assumptions. 

Let $\textbf{Z}=(\textbf{X}, \boldsymbol{y})$ be a sample of size $n$, drawn from some target distribution $p_t(\boldsymbol{x}, y)$. We feed the features $X_i$ within the sample $\textbf{Z}$ through our calibrated model $\textbf{f}$, which results in predictions $\hat{Y}_i$ and calibrated confidence scores $S_i$ with realizations $s_i \in [0,1]$. Let the distribution of confidence scores induced by model $\textbf{f}$ operating on $p_t(\boldsymbol{x}, y)$ be denoted as $p_t(s)$. Now, the number of correct predictions $K_\textbf{Z}$ within the sample $\textbf{Z}$ follows a Poisson binomial distribution with the known expectation $\operatorname{E}[K_\textbf{Z}]=\sum_{i=1}^n S_i$ and the sample accuracy of model $\textbf{f}$ over the sample $\textbf{Z}$ is $\text{Acc}_\textbf{f}(\textbf{Z})=\frac{K_\textbf{Z}}{n}$. This implies that the expected sample accuracy is the sample average of the confidence scores, formally 
\begin{equation*}
\operatorname{E}_{p_t(\boldsymbol{x}, y)}[\text{Acc}_\textbf{f}(\textbf{Z})] = \operatorname{E}_{p_t(\boldsymbol{x}, y)}\left[\frac{K_\textbf{Z}}{n}\right] = \frac{1}{n}\sum_{i=1}^n S_i = \bar{S}_\textbf{Z}.   
\end{equation*}
On the other hand, if we sample a single instance $(X, Y)$ from the target distribution $p_t(\boldsymbol{x}, y)$, we can show the expected value of the corresponding confidence score $S$ to be equal to $\operatorname{Acc}_\textbf{f}(p_t(\boldsymbol{x}, y))$, the accuracy of model $\textbf{f}$ over the whole target distribution.
\begin{lemma}\label{lemma}
    Let $(X,Y)$ be an instance drawn from a target distribution $p_t(\boldsymbol{x}, y)$ and let $(\hat{Y}, S)$ be the corresponding prediction by a calibrated model $\boldsymbol{\mathrm{f}}$. Then,
    \begin{equation*}
        \operatorname{E}_{p_t(s)}[S]=\operatorname{Acc}_{\boldsymbol{\mathrm{f}}}(p_t(\boldsymbol{x}, y)).
    \end{equation*}
\end{lemma}
\begin{proof}
    Recall that $C = g(Y,\hat{Y})$, where $g$ is a (binary-valued) criterion for whether the prediction $\hat{Y}$ is considered to be correct or not. Consider the probability that the prediction was correct given some confidence score $S=s$ and apply the Bayes rule to get
    \begin{equation*}
        P_{p_t(\boldsymbol{x}, y)}(C=1|S=s) = \frac{p_t(s|C=1)P_{p_t(\boldsymbol{x}, y)}(C=1)}{p_t(s)}.
    \end{equation*}
    Using the calibration assumption (Definition~\ref{definition}), we can substitute the left side to get
    \begin{align*}
        s &= \frac{p_t(s|C=1)P_{p_t(\boldsymbol{x}, y)}(C=1)}{p_t(s)} \\
        p_t(s)s &= p_t(s|C=1)P_{p_t(\boldsymbol{x}, y)}(C=1) \\        
        \int_0^1 p_t(s)s~ds &= \int_0^1p_t(s|C=1)P_{p_t(\boldsymbol{x}, y)}(C=1)~ds \\
        \int_0^1 p_t(s)s~ds &= P_{p_t(\boldsymbol{x}, y)}(C=1)\int_0^1p_t(s|C=1)~ds \\
        \operatorname{E}_{p_t(s)}[S] &= P_{p_t(\boldsymbol{x}, y)}(C=1) \\
        \operatorname{E}_{p_t(s)}[S] &= \operatorname{Acc}_{\textbf{f}}(p_t(\boldsymbol{x}, y))
    \end{align*}
\end{proof}
This result can in turn be used to show that the AC estimator is indeed unbiased.

\begin{theorem}\label{unbiased_estimator}
   The sample average of confidence scores $\bar{S}_\textbf{Z}$ is an unbiased estimator for model accuracy $\operatorname{Acc}_{\boldsymbol{\mathrm{f}}}(p_t(\boldsymbol{x}, y))$ in target distribution $p_t(\boldsymbol{x}, y)$. Formally, $\operatorname{E}_{p_t(s)}[\bar{S}_\textbf{Z}]=\operatorname{Acc}_{\boldsymbol{\mathrm{f}}}(p_t(\boldsymbol{x}, y))$. 
\end{theorem}
\begin{proof}
    Using linearity of expectation and Lemma~\ref{lemma}, we have
    \begin{align*}
        \operatorname{E}_{p_t(s)}[\bar{S}_\textbf{Z}] &= \operatorname{E}_{p_t(s)}\left[\frac{1}{n}\sum_{i=1}^n S_i\right] \\
        &= \frac{1}{n}\sum_{i=1}^n \operatorname{E}_{p_t(s)}[S_i] \\
        &= \frac{1}{n}\sum_{i=1}^n \operatorname{Acc}_{\textbf{f}}(p_t(\boldsymbol{x}, y)) \\
        &= \operatorname{Acc}_{\textbf{f}}(p_t(\boldsymbol{x}, y)).
    \end{align*}
\end{proof}
Next, we show that the AC estimator is also consistent.
\begin{theorem}\label{consistent_estimator}
   The sample average of confidence scores $\bar{S}_\textbf{Z}$ is a consistent estimator for sample accuracy $\operatorname{Acc}_{\boldsymbol{\mathrm{f}}}(p_t(\boldsymbol{x}, y))$.
\end{theorem}
\begin{proof}
    By Chebyshev's inequality, it suffices to show that $\operatorname{Var}[\bar{S}_\textbf{Z}]$ tends to zero when $n$ goes to infinity. Since $S_i$ takes a value from the interval $[0,1]$, it is trivial that $\operatorname{E}[S_i^2] \leq \operatorname{E}[S_i]$ for all $i$. Since clearly also $\mu := \operatorname{E}[S_i]\in [0,1]$, it is straightforward to see that the expression $\mu(1-\mu)$ is maximized when $\mu=\frac{1}{2}$, yielding a maximum of $\frac{1}{4}$. Using these insights, we can derive an upper bound for the variance of the estimator as follows.
    \begin{align*}
        \operatorname{Var}[\bar{S}_\textbf{Z}] &= \operatorname{Var}\left[\frac{1}{n}\sum_{i=1}^n S_i\right] \\
        &= \frac{1}{n^2}\sum_{i=1}^n\operatorname{Var}\left[S_i\right] \\
        &= \frac{1}{n^2}\sum_{i=1}^n\left(\operatorname{E}[S_i^2]-\operatorname{E}[S_i]^2\right) \\
        &\leq \frac{1}{n^2}\sum_{i=1}^n\left(\operatorname{E}[S_i]-\operatorname{E}[S_i]^2\right) \\
        &= \frac{1}{n^2}\sum_{i=1}^n\mu(1-\mu) \\
        &\leq \frac{1}{n^2}\sum_{i=1}^n\frac{1}{4} \\
        &= \frac{1}{4n}.
    \end{align*}
    Since $\underset{n\rightarrow\infty}{\lim} \frac{1}{4n} = 0$ and $0 \leq \operatorname{Var}[\bar{S}_\textbf{Z}] \leq \frac{1}{4n}~\forall n$, it necessarily follows that $\underset{n\rightarrow\infty}{\lim} \operatorname{Var}[\bar{S}_\textbf{Z}] = 0$.
    
\end{proof}

Suppose one is only interested in a point estimate for the sample accuracy. In that case, it is sufficient to collect the confidence scores and calculate their mean, justifying the feasibility of AC as an unsupervised accuracy estimator under the calibration assumption. However, one might also be interested in the variance of these estimates. We will deal with this next.

\subsection{Uncertainty in the AC Estimates}

We have shown that under the calibration assumption, AC is an unbiased and consistent estimator of model accuracy, which is the best we can hope for in the sense that even if we could directly measure the model accuracy from a given sample of predictions and their corresponding labels, this would still be only an unbiased and consistent estimate for the model performance on the whole underlying data distribution. However, in the latter case, we would only need to account for the (common cause) variance caused by sampling effects, which in SPC is dealt with control limits. With the AC estimator, there is also uncertainty in the estimates themselves, which adds another layer of variance that one might want to take into account.

\begin{figure*}[ht!]
\centering
\includegraphics[width=1.0\textwidth]{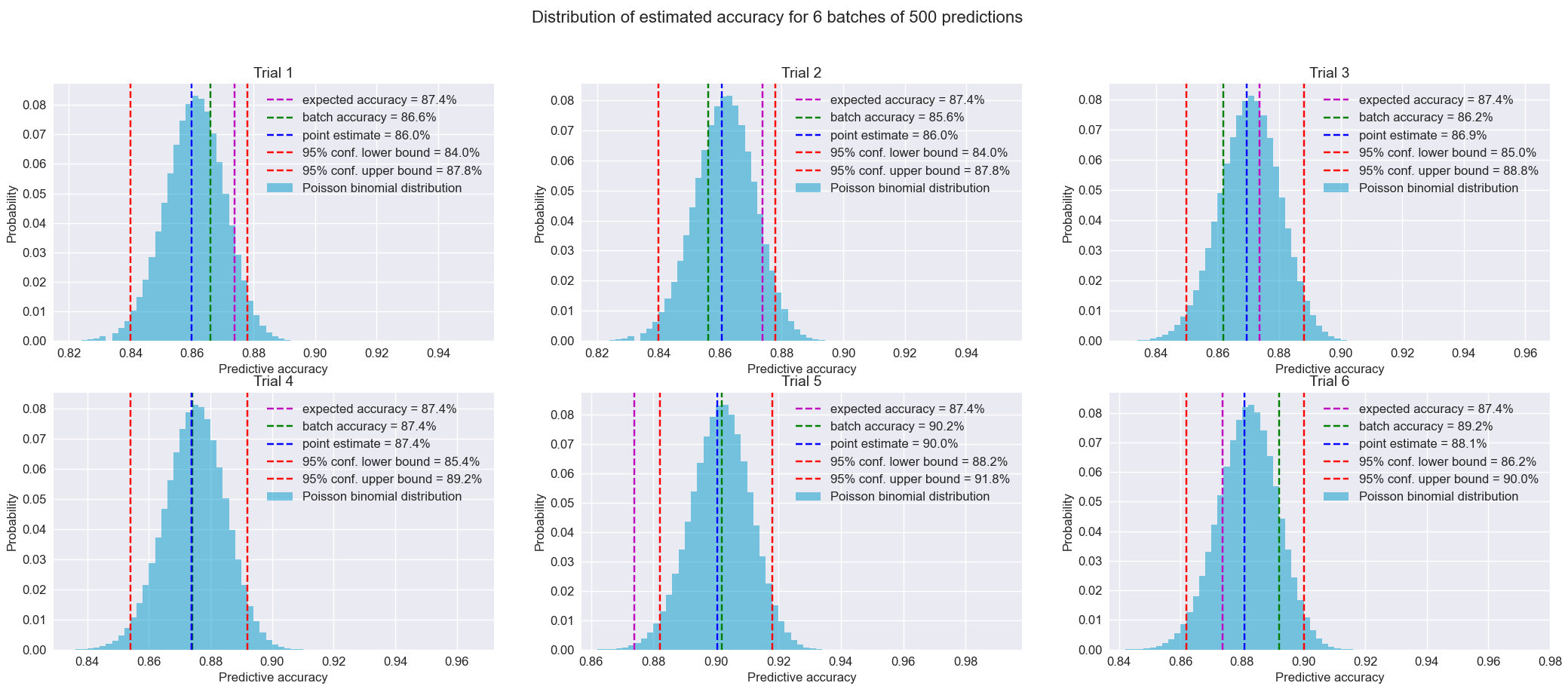}
\caption{An example of estimating the predictive accuracy over 6 batches of 500 predictions using the AC method with CIs. The point estimate (blue line) given by the AC method closely follows the true accuracy (green line) in each batch, which in turn might deviate from the expected accuracy for the whole dataset (magenta line). In each case, the true (batch) accuracy also falls within the predicted 95 \% confidence interval (red lines). The PMF of the Poisson binomial distribution for each batch is shown in light blue.}
\label{PB}
\end{figure*}

Although we might not have access to GT labels to measure sample accuracy, we have already seen that sample accuracy can be modeled as a random variable following a Poisson binomial distribution with its expected value equal to the sample average of confidence scores. Thus, one can choose to use the CDF of the Poisson binomial distribution to estimate confidence intervals (CI) for the sample accuracy. For example, one can estimate the (central) $95\%$ CI for the sample accuracy $\operatorname{Acc}_\textbf{f}(\textbf{Z})$ in a given sample $\textbf{Z}$ by first deriving the CDF for the Poisson binomial distribution of $\frac{K_\textbf{Z}}{n}$ and then finding the range where $0.025 \leq F\left(\frac{K_\textbf{Z}}{n}\right) \leq 0.975$. We will empirically demonstrate this choice's effectiveness in Chapter~\ref{experiments}. An illustration of the AC method with CIs is given in Fig.~\ref{PB}

\subsection{Estimating the Confusion Matrix for Failure Prediction}\label{confusion}

Sometimes confidence scores are used as a tool for \textit{failure prediction}~\shortcite{hendrycks:2017,kivimaki:2023}. A failure prediction mechanism based on confidence scores is a mapping $h:S\rightarrow\{0,1\}$. It sets a \textit{confidence threshold} $t \in [0, 1]$ and treats all predictions with a confidence score equal to or above the threshold as correct and below the threshold as incorrect. Predictions with sub-threshold confidence can be altogether discarded, or sent to further inspection by some downstream system or human-in-the-loop. This might be useful, especially in situations where different types of errors induce different costs. It is worthwhile to stress that this treatment creates second-order predictions (predictions about predictions) and is not to be confused with the true underlying discriminator $g$. 

Suppose such a failure prediction mechanism is applied. In that case, one can leverage the associated (binary) confusion matrix, which collects the numbers of True Positives ($TP$), False Positives ($FP$), True Negatives ($TN$), and False Negatives ($FN$), in monitoring the mechanism.  Again, if GT is not available, one can only estimate these quantities. If the calibration assumption holds, we can use the expected values for all of the elements of the confusion matrix as point estimates. This gives us the estimated confusion matrix shown in Table~\ref{table: confusion}. 

\begin{table}[ht!]
\small
\centering
\begin{tabular}{rr||c|c} 
& & \multicolumn{2}{c}{Predicted}\\
 & & Correct & Incorrect \\ 
\hline \hline
\parbox[t]{0mm}{\multirow{2}{*}{\rotatebox[origin=c]{90}{Actual~~}}} & 
Correct & \(\displaystyle TP = \sum_{S_i \geq t} S_i\) & \(\displaystyle FN = \sum_{S_i < t} S_i\)  \\
& Incorrect & \(\displaystyle FP = \sum_{S_i \geq t} (1-S_i)\) & \(\displaystyle TN = \sum_{S_i < t} (1-S_i)\) \\
\hline
\end{tabular}
\caption{Estimated confusion matrix}
\label{table: confusion}
\end{table}

The estimated confusion matrix can be further used to derive estimates for different performance metrics, such as precision, recall, F$_1$, and accuracy. If the user wishes to estimate the performance over all possible values of $t$, an ROC curve can be plotted and the area under the ROC curve (AUROC) can be used as a general performance estimate. It is important to remember that all of these metrics estimate the performance of the failure prediction mechanism and not the performance of the base model. For example, the estimated predictive accuracy of the base model can be decomposed as
\begin{equation*}
    \frac{1}{n}\sum_{i=1}^n S_i = \frac{1}{n}\left(\sum_{S_i<t} S_i + \sum_{S_i \geq t} S_i\right) = \frac{1}{n}(TP+FN),
\end{equation*}
whereas the estimated accuracy of the failure prediction mechanism is $\frac{1}{n}(TP+TN)$. These different types of accuracies should not be confused with each other.

\section{Experiments}\label{experiments}

We conduct two experiments\footnote{Code is publicly available at \url{https://github.com/JuhaniK/AC_trials}} with synthetic data. In the first experiment, we examine the quality of point predictions and confidence intervals of the estimates derived using the AC method when the confidence scores are known to be calibrated. In the second experiment, we compare AC, DoC-Feat, and ATC under covariate shift when used to estimate predictive accuracy in a monitoring setting.

\subsection{Experimenting with Calibrated Data}

Since with any real-life data, the inevitable calibration error will affect the outcome of our experiments, we will first create a simulated dataset, which will be calibrated by construction. We will begin by describing the data creation process.

\subsubsection{Data Description}\label{simulation}

As stated in Section~\ref{background}, calibration error is caused by the confidence scores not aligning with empirical probabilities. We can try to minimize this discrepancy by first creating a set of confidence scores. Then, for each element in the set of confidence scores, we can sample a binary label from a Bernoulli distribution using the confidence score as the parameter. This way of sampling ensures that the confidence scores will align with empirical probabilities, albeit some small amount of calibration error might persist due to sampling effects. However, it will suffice to validate the properties discussed in Section~\ref{method} empirically.

We chose to create our simulated set of confidence scores by drawing samples from a mixture of three Beta distributions. The first component was biased to generate high-confidence predictions, the second to generate average ones, and the third to create low-confidence predictions. By adjusting the mixture weights, we could effectively emulate distributions of confidence scores outputted by models operating on real-life data. Then, by changing the mixture weights, we could easily simulate shifts in model performance to see how the estimator under scrutiny managed to keep track of them. The original dataset was trying to emulate a typical distribution of confidence scores where most predictions are made with high confidence~\shortcite{guo:2017}. For the shifted distribution, the fraction of high-confidence predictions was slightly decreased. The exact parameters and weights of the mixture components used in the experiment are given in Table~\ref{table: components}.

\begin{table}[ht!]
\small
\centering
\begin{tabular}{r|c|c|c|c} 
& \multicolumn{2}{c|}{Parameters} & \multicolumn{2}{c}{Weights} \\
Components & $\alpha$ & $\beta$ & Original & Shifted\\ 
\hline 
Component 1 & 20 & 1 & 0.9 & 0.8 \\
Component 2 & 2 & 2 & 0.08 & 0.15\\
Component 3 & 1 & 20 & 0.02 & 0.05 \\ 
\hline
\end{tabular}
\caption{Parameters and weights of the Beta mixture components}
\label{table: components}
\end{table}

\begin{figure}[ht!]
\centering
\includegraphics[width=0.84\textwidth]{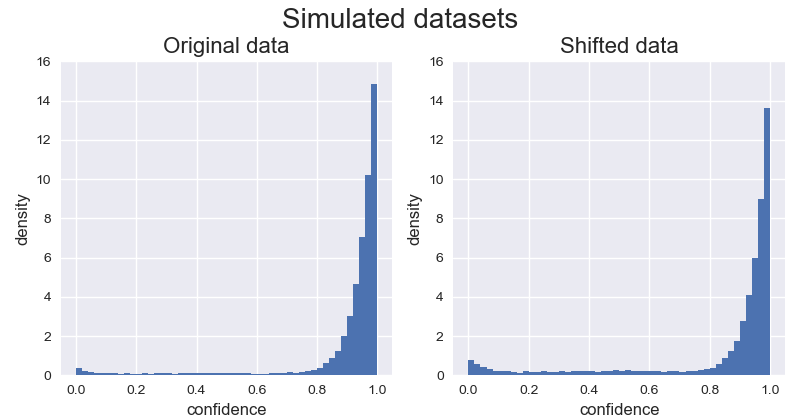}
\caption{Distribution of confidence scores in the simulated data.}
\label{simulated}
\end{figure}

The distribution of confidence scores for these two distributions is sketched in Fig.~\ref{simulated}. The overall accuracy for the original dataset is $89.7\%$ and $83.8\%$ for the shifted dataset. Although both datasets should be perfectly calibrated by construction, they both express slight calibration errors due to sampling effects as discussed above. The calibration errors (ACE) are $0.63\%$ for the original and $0.64\%$ for the shifted dataset respectively.

\subsubsection{Quality of Point Estimates under Shift}

\begin{figure}[ht!]
\centering
\includegraphics[width=0.84\textwidth]{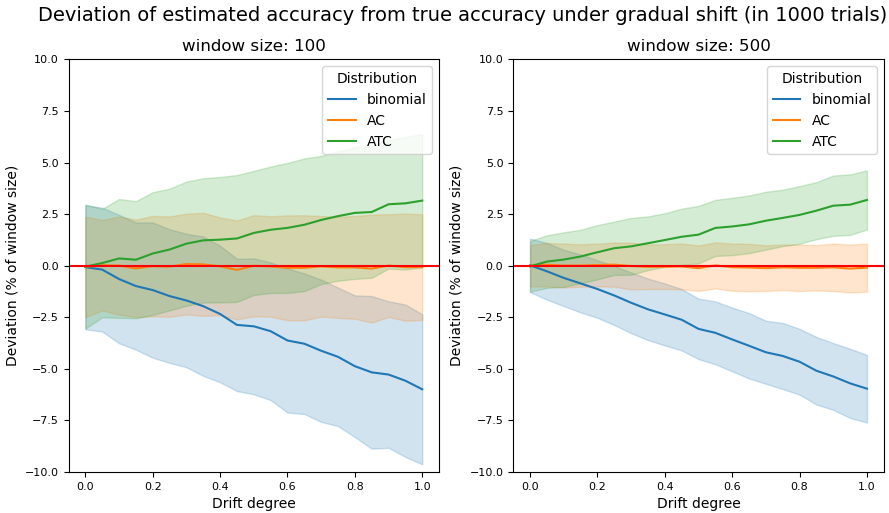}
\caption{The quality of point estimates under gradual data drift.}
\label{pe}
\end{figure}

We tested the quality of point estimates for predictive accuracy derived using AC by simulating gradual data shifts. We used two different monitoring window sizes, namely 100 and 500 predictions. We gradually increased the fraction of instances sampled from the shifted distributions in increments of $5\%$, starting from $0\%$ and ending in $100\%$. For each degree of shift, we conducted 1,000 trials, where we measured the error of the estimate in each of the trials and plotted the mean of the errors in Fig.~\ref{pe}, along with one standard deviation (shaded area). We contrasted the results of AC by also similarly plotting estimates derived by using either ATC~\shortcite{garg:2022} or a binomial assumption (using the overall accuracy of the original distribution as the parameter). The binomial assumption comparison was included to give a sense of how fast the estimation error would increase if the predictive performance was assumed not to change. 

We excluded DoC-Feat from this comparison, since under the calibration assumption, DoC-Feat is equal to AC. In fact, it is straightforward to verify that Doc-Feat can improve on AC only, if the model is either over- or underconfident to begin with and the predictions become even more over- or underconfident. In cases where an underconfident model becomes overconfident after the shift, or vice versa, or if the discrepancy between the AC estimate and true accuracy gets smaller, DoC-Feat will necessarily yield worse estimates than AC.  

The results show that AC yields point estimates, which on average align with the true sample accuracy. In contrast, ATC underestimates the true accuracy systematically with increasing shifts. Unsurprisingly, the binomial assumption of no change in the predictive performance leads to overestimating the true accuracy. In both cases, the AC method also yields smaller variances in the estimates, with the variance decreasing with a larger window size.

\subsubsection{Quality of Confidence Intervals}

We tested the quality of the estimated confidence intervals of the AC method by first estimating the $95\%$ CI as described in Section~\ref{method} and checking how often the actual sample accuracy fits inside the estimated interval. We did this for both the original and shifted data and for window sizes of $\{100, 200, 300, 400, 500\}$. In each case, we conducted 10,000 trials and calculated the fraction of times the actual accuracy was within the estimated CI. For good quality estimates this fraction should be close to $95\%$. This generally desired property of CIs is referred to as \textit{validity}. Another desired property is \textit{optimality}, which states that the method used in constructing the CIs should use as much of the information in the dataset as possible.

\begin{figure}[t!]
\centering
\includegraphics[width=0.92\textwidth]{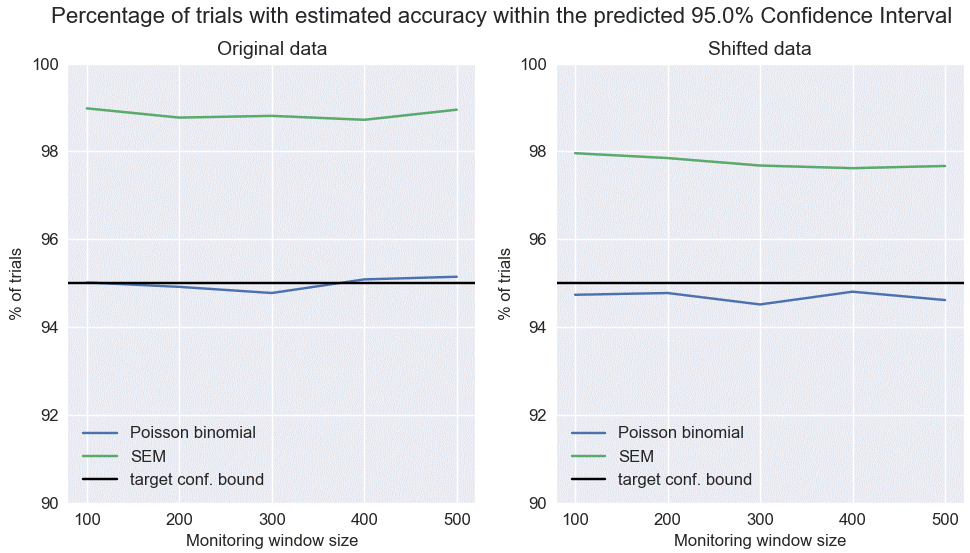}
\caption{The quality of the estimated CIs for the original and shifted data. }
\label{ci}
\end{figure}

Since none of the other methods discussed in Section~\ref{related} yield confidence intervals for their estimates, we contrasted the CIs of AC derived using the properties of the Poisson binomial distribution against the method used in NannyML\footnote{https://www.nannyml.com/}, which is an open-source library for model monitoring. NannyML implements a performance estimation algorithm, namely Confidence-based Performance Estimator (CBPE), which uses calibrated confidence scores to estimate the confusion matrix similarly to what was described in Section~\ref{confusion}. Estimates for all typical classification metrics such as precision, recall, F$_1$, AUROC, and so on can then be derived from this matrix. Notably, when CBPE is used to estimate model accuracy, it results in point estimates identical to AC. However, it has a different way of estimating confidence bands. In CPBE, the confidence band is defined as Sampling Error, which is set to be $\pm 3$ Standard Errors of the Mean (SEM) $\sigma_{\bar{x}}$, which is calculated as
\begin{equation}
    \sigma_{\bar{x}} = \frac{\sigma}{\sqrt{n}},
\end{equation}
where $\sigma$ is an estimation of the population variance as the variance of the true labels in the reference data set (where GT is known) and $n$ is the monitoring window size (NannyML uses the term "chunk size"). We replaced the $\pm 3$ standard errors with $\pm 1.96$ standard errors in our tests to get the lower and upper $95\%$ confidence limits for a fair comparison against the CIs derived using the Poisson binomial approach. The results of this experiment are presented in Fig.~\ref{ci}, clearly showing the superiority of the Poisson binomial approach, which produces (roughly) valid CIs. In contrast, the confidence bands produced by the SEM approach are extremely conservative. 

Using SEM to construct confidence bands only accounts for the uncertainty caused by the sample accuracies of the chunks not aligning with the (potentially shifted) distribution accuracy due to sampling effects. The uncertainty in the estimates for the sample accuracy is omitted, which implicitly equates to treating the mean of confidence scores in a batch as if it matched the sample accuracy exactly. In fact, the only information used from each batch is the chunk size, which results in fixed confidence bands for each batch. Furthermore, the SEM itself can only be estimated from the reference data since the population variance is not known, and since the required i.i.d. assumption gets broken under distributional shift, the estimate becomes an approximation. 

We argue that it is conceptually clearer to account for the variance due to sampling effects with control limits and form the confidence bands for the estimated accuracy solely on the basis of uncertainty in the estimates themselves using the Poisson binomial approach. In this way, all available information from each batch gets utilized and the width of the confidence band is allowed to vary from batch to batch. We are currently working to implement this improved approach to forming confidence bands into the NannyML library.

\subsection{Experimenting under Covariate Shift}\label{covariate_shift}

Next, we compared the confidence-based accuracy estimators AC, DoC-feat, and ACT with both uncalibrated and calibrated confidence scores. Earlier studies~\shortcite{guillory:2021,garg:2022} have had conflicting results on whether calibration has a beneficial effect in estimating predictive accuracy or not. In these studies, accuracy estimation has been performed on the dataset level (with sample size in the order of thousands or tens of thousands), whereas we are interested in estimates derived for sample sizes generally used in model monitoring (in the order of hundreds). Furthermore, they have emphasized neural networks used in image classification, with the nature of shifts in the shifted datasets not fully controlled. We expanded this scope by comparing the estimation methods using seven non-neural classifier models, namely Logistic Regression (LR), (Gaussian) Naïve Bayes (NB), K-Nearest Neighbors (KNN), Support Vector Machine (SVM) with RBF kernel, and Random Forest (RF), using implementations provided by the scikit-learn library~\footnote{https://scikit-learn.org/stable/}, as well as XGBoost (XGB)~\footnote{https://xgboost.readthedocs.io/en/stable/}, and Light-GBM (LGBM)~\footnote{https://lightgbm.readthedocs.io/en/latest/index.html} to see whether the earlier results would generalize to these models as well. All models were trained with the default parameter settings of their respective implementations. 

\subsubsection{Data and Model Training Descriptions}

We used simulated data in order to have full control over the nature of the shift. Since all confidence-based accuracy estimators will eventually fail under concept shift~\shortcite{garg:2022}, we restricted the shift as covariate shift. We experimented with two scenarios, where in both scenarios our data consisted of two features and a binary label (1 or 0). 

In both scenarios, we drew samples from a mixture of two-dimensional Gaussians, where the modes would form a symmetrical pattern. In the first scenario, we assigned label values for each datapoint stochastically, according to their distance from the line $y=x$. Points falling exactly on the line had a 50\% probability for either label. For data points sampled from modes further from the line $y=x$, the majority of samples would be labeled as either 0 or 1 depending on which side of the line they fell, making them easy to predict. Points sampled from modes closer to the line would have higher label entropy, making them hard to predict. In the second scenario, the label values would depend on their distance from the circle $x^2+y^2=5$, where points falling on the circle had a 100\% probability of being assigned label 1. This probability decreased rapidly as points moved further from the circle (either inside or outside). This time, points falling close to the circle or far from it could be considered easy and points falling somewhere in between could be considered hard. A more detailed description of the data generation process is given in Appendix A, along with some visualizations.

These design choices enabled two properties. First, they made it relatively easy for each of the trained models to find a decision boundary, which more or less aligned with the Bayes optimal decision boundary, at least in the first scenario. Second, we could alter the portions of easy and hard samples to create covariate shift, which altered the prediction accuracy of the models without inducing concept shift. In the first scenario, we could examine the performance of the different estimators with different models when the Bayes optimal decision boundary was linear, and in the second scenario when it was not.

In both scenarios, we trained the models with 100,000 samples, of which $80\%$ were easy and $20\%$ were hard. Using the same ratio, we drew 25,000 additional samples to train the calibration mappings for each model and 25,000 additional samples for the setup required by the DOC-Feat and ATC methods. Since there was no shortage of data, we used the non-parametric Isotonic Regression~\shortcite{zadrozny:2002} to derive the calibration mappings. It is generally considered to produce good quality calibration mappings but is also known to sometimes overfit with smaller datasets~\shortcite{kull:2017}. In each trial, we created a test dataset of 25,000 samples, with an increasing portion of hard-to-predict samples, as described in Table~\ref{portions}.
\begin{table}[!ht]
    \centering
    \begin{tabular}{|c|c|c|} \hline
        Shift & Easy & Hard  \\ \hline
        0 & 20,000 & 5,000  \\ 
        1 & 15,000 & 10,000  \\ 
        2 & 12,500 & 12,500  \\ 
        3 & 10,000 & 15,000  \\      
        \hline
    \end{tabular}
    \caption{The number of easy and hard samples for each degree of shift}
    \label{portions}
\end{table}

The accuracy and calibration error for the Bayes optimal classifier are given in Table~\ref{bayes}, which provides a useful point of reference for the results to follow. We see that in both scenarios, the covariate shift induced a roughly similar gradual decrease in accuracy, although the decision boundary did not change. It is to be noted that the calibration error for the Bayes optimal classifier is only due to sampling effects and increasing the sample size to infinity would result in zero calibration error. 
\begin{table}[!ht]
    \centering
    \begin{tabular}{|c|c|c|c|c|} \hline
        ~ & \multicolumn{2}{c|}{Linear boundary} & \multicolumn{2}{c|}{Non-linear boundary}\\ \hline
        Shift & Accuracy & ACE & Accuracy & ACE \\ \hline
        0 & 91.7\% & 0.5\% & 92.3\% & 0.5\% \\ 
        1 & 90.3\% & 0.5\% & 89.9\% & 0.5\% \\ 
        2 & 90.0\% & 0.7\% & 88.7\% & 0.9\% \\ 
        3 & 88.7\% & 0.5\% & 87.1\% & 0.8\% \\ \hline
    \end{tabular}
    \caption{Accuracy and calibration error of the Bayes optimal classifier for both scenarios.}
    \label{bayes}
\end{table}

For both scenarios, the experiment was performed as follows: In order to simulate a monitor deriving estimates from a limited amount of samples, we performed 1,000 trials for each degree of shift. In each trial, we drew 500 samples with replacement from the sampled test dataset as a simulated batch of incoming data. For each model, we report the mean of sample accuracies over all trials. We also report the calibration error before and after applying the calibration mapping, calculated for the whole dataset in each case. We calculated the estimated accuracy for each estimator and compared that against the true sample accuracy. For each estimator, we report the mean of absolute estimation error over all trials. 

\subsubsection{Results with Linear Decision Boundary}

The results summarized in Table~\ref{gradient_shift} show that none of the estimators can be considered a clear winner. In all cases, the difference in mean absolute estimation errors for the top-performing estimators is relatively small. In comparison, the standard deviations for different models and degrees of shift were typically in the range of $1.2\%-2.0\%$ over the 1,000 trials. Overall, AC and DoC-Feat variants seem to produce comparable estimates, with AC$_c$ having a slight overall advantage.
\setlength{\tabcolsep}{5pt}
\begin{table*}[ht]    
    \centering
 \begin{tabular}{|c|c|c|c|c|c|c|c|c|c|c|}
    \hline
        Shift & Model & Acc & ACE$_u$ & ACE$_c$ & AC$_u$ & AC$_c$ & DOC$_u$ & DOC$_c$ & ATC$_u$ & ATC$_c$ \\ \hline
       \multirow{7}{*}{0} & NB & 91.7\% & 4.3\% & 0.4\% & 3.9\% & \textbf{1.0\%} & 8.5\% & 1.3\% & 1.4\% & 1.4\% \\ 
         & LR & 91.7\% & 3.1\% & 0.4\% & 1.5\% & \textbf{1.0\%} & 3.5\% & 1.3\% & 1.4\% & 1.4\% \\ 
         & KNN & 91.1\% & 7.4\% & 0.6\% & 1.6\% & \textbf{1.0\%} & \textbf{1.0\%} & 1.3\% & 22.4\% & 9.7\% \\ 
         & SVM & 91.6\% & 2.4\% & 0.3\% & \textbf{1.0\%} & \textbf{1.0\%} & 1.5\% & 1.3\% & 1.5\% & 1.3\% \\ 
         & RF & 90.2\% & 5.9\% & 0.6\% & 2.4\% & \textbf{1.0\%} & 1.1\% & 1.1\% & 1.3\% & 1.3\% \\ 
         & XGB & 91.7\% & 0.8\% & 0.5\% & \textbf{1.0\%} & \textbf{1.0\%} & 1.3\% & 1.3\% & 1.3\% & 1.3\% \\ 
         & LGBM & 91.7\% & 0.6\% & 0.5\% & \textbf{1.0\%} & \textbf{1.0\%} & 1.3\% & 1.3\% & 1.4\% & 1.3\% \\ 
\hline
       \multirow{7}{*}{1}
         & NB & 90.3\% & 5.8\% & 0.7\% & 5.6\% & \textbf{1.1\%} & 10.2\% & 1.4\% & 3.3\% & 3.3\% \\ 
         & LR & 90.3\% & 3.8\% & 0.6\% & 2.9\% & \textbf{1.0\%} & 4.9\% & 1.4\% & 3.3\% & 3.3\% \\ 
         & KNN & 89.4\% & 8.6\% & 1.3\% & 1.7\% & 1.4\% & \textbf{1.0\%} & 1.7\% & 25.6\% & 12.1\% \\ 
         & SVM & 90.3\% & 2.1\% & 0.6\% & \textbf{1.2\%} & \textbf{1.2\%} & 1.7\% & 1.5\% & 2.9\% & 4.0\% \\ 
         & RF & 88.2\% & 5.4\% & 2.0\% & 2.9\% & 1.8\% & \textbf{1.4\%} & 1.9\% & \textbf{1.4\%} & 1.6\% \\ 
         & XGB & 90.2\% & 0.5\% & 0.8\% & \textbf{1.0\%} & \textbf{1.0\%} & 1.3\% & 1.4\% & 3.2\% & 3.3\% \\ 
         & LGBM & 90.2\% & 0.6\% & 0.6\% & \textbf{1.0\%} & \textbf{1.0\%} & 1.3\% & 1.3\% & 3.1\% & 3.4\% \\ 
\hline
       \multirow{7}{*}{2}
         & NB & 90.0\% & 6.6\% & 0.7\% & 6.6\% & \textbf{1.2\%} & 11.2\% & 1.5\% & 5.0\% & 5.1\% \\ 
         & LR & 90.0\% & 4.5\% & 0.8\% & 3.8\% & \textbf{1.1\%} & 5.8\% & 1.5\% & 5.1\% & 5.2\% \\ 
         & KNN & 88.9\% & 6.3\% & 1.4\% & 1.7\% & 1.7\% & \textbf{1.1\%} & 2.0\% & 26.9\% & 12.6\% \\ 
         & SVM & 90.0\% & 2.6\% & 0.6\% & 1.2\% & \textbf{1.1\%} & 1.7\% & 1.5\% & 4.2\% & 5.5\% \\ 
         & RF & 87.6\% & 5.3\% & 2.1\% & 2.8\% & 2.0\% & \textbf{1.3\%} & 2.1\% & 1.7\% & 2.2\% \\ 
         & XGB & 89.9\% & 0.9\% & 0.6\% & \textbf{1.0\%} & \textbf{1.0\%} & 1.3\% & 1.4\% & 4.6\% & 4.7\% \\ 
         & LGBM & 89.9\% & 0.7\% & 0.6\% & \textbf{1.1\%} & \textbf{1.1\%} & 1.4\% & 1.4\% & 4.9\% & 5.1\% \\ 
\hline
       \multirow{7}{*}{3}
         & NB & 88.7\% & 6.9\% & 0.5\% & 6.8\% & \textbf{1.1\%} & 11.4\% & 1.4\% & 6.4\% & 6.6\% \\ 
         & LR & 88.7\% & 4.7\% & 0.6\% & 3.9\% & \textbf{1.1\%} & 5.9\% & 1.5\% & 6.4\% & 6.7\% \\ 
         & KNN & 87.5\% & 8.3\% & 2.8\% & 2.2\% & 2.6\% & \textbf{1.5\%} & 2.9\% & 28.1\% & 13.2\% \\ 
         & SVM & 88.7\% & 3.1\% & 1.3\% & 1.8\% & \textbf{1.5\%} & 2.3\% & 1.9\% & 5.4\% & 6.7\% \\ 
         & RF & 86.4\% & 5.7\% & 3.0\% & 3.5\% & 2.9\% & 1.8\% & 2.9\% & \textbf{1.6\%} & 1.9\% \\ 
         & XGB & 88.6\% & 0.9\% & 0.9\% & \textbf{1.1\%} & \textbf{1.1\%} & 1.3\% & 1.5\% & 6.1\% & 6.2\% \\ 
         & LGBM & 88.6\% & 0.6\% & 0.6\% & \textbf{1.1\%} & \textbf{1.1\%} & 1.4\% & 1.4\% & 5.8\% & 6.2\% \\ \hline
         & Mean & 89.7\% & 3.9\% & 0.9\% & 2.4\% & \textbf{1.3}\% & 3.2\% & 1.6\% & 6.6\% & 4.9\% \\ \hline
    \end{tabular}
    \caption{Mean absolute error of estimated accuracy using data with linear decision boundary. "Acc" stands for Accuracy, "ACE" for Adaptive Expected Calibration Error, "AC" for Average Confidence, "DOC" for DoC-Feat, and "ATC" for Average Thresholded Confidence. Underscipts $u$ and $c$ refer to using uncalibrated or calibrated confidence scores respectively. All values are the averages over the 1,000 trials, except for calibration errors, which are calculated on the dataset level. The best estimator for each model and degree of shift is bolded.}
    \label{gradient_shift}
\end{table*}

Except for RF, the ATC variants yield notably higher estimation errors. For the AC method, using calibrated confidence scores tends to yield better results, except for XGB and LGBM, which express low calibration errors for all levels of shift even when uncalibrated. With DoC-Feat, using calibrated confidence scores results in better quality estimates when using NB, LR, or SMV, and equal or worse estimates with every other model. For ATC, calibration does not seem to yield a significant effect, except for KNN, which benefits from calibration.

In all cases, using either variant of ATC results in huge estimation errors, when applied to KNN. This might result from the way the probabilities are predicted with KNN as voting ratios, which generates coarse-grained confidence scores overfitting to noisy data. Increasing the number of nearest neighbors $k$ might help to counterbalance this. 

It is worthwhile to notice that the differences between the estimates for different models do not depend on the differences between their predictive accuracies. All models roughly match the performance of the Bayes optimal classifier, except for RF and KNN, which fall behind only slightly. For XGB and LGBM, applying a calibration mapping does not have a significant effect on calibration error. For all the other models, applying a calibration has a clear beneficial effect, which persists over the shifts, although the calibration error does slightly increase for all models. 

Interestingly, there seems to be a strong positive correlation between estimation error and calibration error for AC$_c$, and DoC-Feat$_c$. We illustrate this in Table~\ref{correlation}, where estimators using uncalibrated confidence scores are compared against ACE$_u$, and estimators using calibrated confidence scores are compared against ACE$_c$. Correlation is measured as the Pearson correlation coefficient over all models and degrees of shift using the mean absolute errors as estimates. 
\begin{table}[!ht]
    \centering
    \begin{tabular}{c|c|c} 
        ~ & \multicolumn{2}{c}{Correlation} \\ \hline
        Estimator & Linear boundary & Non-linear boundary \\ \hline
        AC$_u$ & 0.580 & 0.918 \\
        AC$_c$ & 0.980 & 0.991 \\
        DOC-Feat$_u$ & 0.338 & 0.904 \\
        DOC-Feat$_c$ & 0.954 & 0.991 \\
        ATC$_u$ & 0.580 & 0.699 \\
        ATC$_c$ & 0.263 & 0.611 \\ \hline
    \end{tabular}
    \caption{The Pearson correlation coefficients between estimation and calibration errors for different estimators.}
    \label{correlation}
\end{table}

\subsubsection{Results with Non-linear Decision Boundary}

The results summarized in Table~\ref{circle_shift} again show no single best estimator, although AC$_c$ and DOC-Feat$_c$ have the best average performance. For LR and NB, which have limited complexity, the performance of all estimators is rendered unusable even under a slight shift. For all other models, the variants of AC and DOC-Feat yield the best estimates, except for RF, where ATC$_u$ yields comparable results. With both AC and DoC-Feat methods, calibration seems to be beneficial for all but SVM, XGB, and LGBM models, where the differences between calibrated and uncalibrated versions are negligible. For ATC, applying calibration leads to worse performance in almost all cases. 

\begin{table*}[!ht]
    \centering
    \begin{tabular}{|c|c|c|c|c|c|c|c|c|c|c|}
    \hline
        Shift & Model & Acc & ACE$_u$ & ACE$_c$ & AC$_u$ & AC$_c$ & DOC$_u$ & DOC$_c$ & ATC$_u$ & ATC$_c$ \\ 
   	\hline
	\multirow{7}{*}{0}
         & NB & 80.1\% & 21.4\% & 0.4\% & 2.1\% & \textbf{1.2\%} & \textbf{1.2\%} & \textbf{1.2\%} & 2.0\% & 14.8\% \\ 
         & LR & 80.1\% & 13.2\% & 0.1\% & 1.4\% & 1.4\% & 1.4\% & \textbf{1.3\%} & 2.0\% & 64.6\% \\ 
         & KNN & 91.3\% & 5.4\% & 1.1\% & 2.4\% & \textbf{0.9\%} & \textbf{0.9\%} & 1.0\% & 14.7\% & 4.4\% \\ 
         & SVM & 92.0\% & 2.1\% & 0.6\% & \textbf{0.9\%} & \textbf{0.9\%} & 1.3\% & \textbf{0.9\%} & 1.1\% & 1.1\% \\ 
         & RF & 90.6\% & 6.0\% & 1.1\% & 3.2\% & \textbf{0.9\%} & \textbf{0.9\%} & 1.0\% & 1.1\% & 1.2\% \\ 
         & XGB & 92.0\% & 0.4\% & 0.4\% & \textbf{0.9\%} & \textbf{0.9\%} & \textbf{0.9\%} & 1.1\% & 1.1\% & 1.1\% \\ 
         & LGBM & 92.2\% & 0.4\% & 0.5\% & \textbf{0.9\%} & \textbf{0.9\%} & 1.1\% & 1.1\% & 1.1\% & 1.3\% \\ 

	\hline
	\multirow{7}{*}{1}
         & NB & 65.8\% & 29.1\% & 8.7\% & 14.7\% & \textbf{7.9\%} & 12.8\% & \textbf{7.9\%} & 14.4\% & 14.5\% \\ 
         & LR & 65.8\% & 22.1\% & 13.9\% & 14.3\% & \textbf{13.6\%} & 14.3\% & \textbf{13.6\%} & 16.1\% & 54.0\% \\ 
         & KNN & 88.7\% & 3.1\% & 2.3\% & 3.0\% & \textbf{1.1\%} & 1.2\% & \textbf{1.1\%} & 18.6\% & 9.3\% \\ 
         & SVM & 89.6\% & 3.6\% & 0.9\% & 1.1\% & \textbf{1.0\%} & 1.5\% & \textbf{1.0\%} & 1.3\% & 2.5\% \\ 
         & RF & 87.8\% & 6.3\% & 2.7\% & 4.2\% & \textbf{1.1\%} & 1.5\% & 1.2\% & 1.4\% & 2.2\% \\ 
         & XGB & 89.8\% & 0.6\% & 0.7\% & \textbf{1.0\%} & \textbf{1.0\%} & \textbf{1.0\%} & 1.2\% & 1.6\% & 2.2\% \\ 
         & LGBM & 89.8\% & 0.7\% & 0.6\% & \textbf{1.0\%} & \textbf{1.0\%} & 1.2\% & 1.2\% & 1.4\% & 2.6\% \\ 

	\hline
	\multirow{7}{*}{2}
         & NB & 59.4\% & 32.9\% & 12.1\% & 20.3\% & \textbf{11.0\%} & 18.4\% & 11.1\% & 20.7\% & 14.7\% \\ 
         & LR & 59.4\% & 27.1\% & 20.1\% & 20.8\% & 19.8\% & 20.8\% & \textbf{19.7\%} & 23.6\% & 49.3\% \\ 
         & KNN & 87.1\% & 3.9\% & 2.4\% & 3.3\% & \textbf{1.2\%} & 1.5\% & \textbf{1.2\%} & 20.9\% & 11.9\% \\ 
         & SVM & 87.8\% & 4.8\% & 1.4\% & \textbf{1.1\%} & 1.2\% & 1.6\% & 1.2\% & 1.4\% & 3.3\% \\ 
         & RF & 86.2\% & 6.5\% & 3.0\% & 4.7\% & \textbf{1.2\%} & 2.0\% & 1.3\% & 1.5\% & 2.9\% \\ 
         & XGB & 88.4\% & 1.1\% & 0.7\% & \textbf{1.0\%} & 1.1\% & \textbf{1.0\%} & 1.3\% & 2.3\% & 3.5\% \\ 
         & LGBM & 88.5\% & 1.1\% & 0.9\% & 1.1\% & \textbf{1.0\%} & 1.3\% & 1.2\% & 2.0\% & 4.1\% \\ 

	\hline
	\multirow{7}{*}{3}
         & NB & 51.8\% & 36.6\% & 16.2\% & 27.2\% & 15.4\% & 25.2\% & 15.4\% & 28.2\% & \textbf{14.2\%} \\ 
         & LR & 51.8\% & 32.1\% & 27.2\% & 28.2\% & 26.8\% & 28.2\% & \textbf{26.7\%} & 32.1\% & 43.9\% \\ 
         & KNN & 85.5\% & 5.0\% & 3.4\% & 4.0\% & \textbf{1.4\%} & 2.0\% & \textbf{1.4\%} & 22.2\% & 13.9\% \\ 
         & SVM & 86.3\% & 5.2\% & 1.0\% & \textbf{1.2\%} & 1.4\% & 1.6\% & 1.4\% & 1.5\% & 4.0\% \\ 
         & RF & 84.4\% & 7.8\% & 4.1\% & 5.7\% & \textbf{1.5\%} & 2.9\% & 1.6\% & \textbf{1.5\%} & 3.0\% \\ 
         & XGB & 87.0\% & 1.0\% & 0.8\% & \textbf{1.0\%} & \textbf{1.0\%} & \textbf{1.0\%} & 1.3\% & 2.5\% & 4.1\% \\ 
         & LGBM & 86.9\% & 1.2\% & 1.1\% & \textbf{1.1\%} & \textbf{1.1\%} & 1.3\% & 1.3\% & 2.0\% & 4.5\% \\ \hline
         & Mean & 81.6\% & 10.0\% & 4.6\% & 6.1\% & \textbf{4.2\%} & 5.4\% & 4.3\% & 8.6\% & 12.6\% \\ \hline
    \end{tabular}
    \caption{Mean absolute error of estimated accuracy using data with non-linear decision boundary. "Acc" stands for Accuracy, "ACE" for Adaptive Expected Calibration Error, "AC" for Average Confidence, "DOC" for DoC-Feat, and "ATC" for Average Thresholded Confidence. Underscipts $u$ and $c$ refer to using uncalibrated or calibrated confidence scores respectively. All values are the averages over the 1,000 trials, except for calibration errors, which are calculated on the dataset level. The best estimator for each model and degree of shift is bolded.}
    \label{circle_shift}
\end{table*}

In this second scenario, NB and LR are rendered altogether useless. Their predictive power reduces so rapidly under shift that at degree 3 they are essentially reduced to a coin toss. Although their calibration error can be reduced close to zero in the no-shift setting by applying a calibration mapping, the calibration errors for both the uncalibrated and calibrated confidence scores go through the roof under shift. 

The accuracies of all other models are relatively on par and close to the accuracy of the Bayes optimal classifier, with RF and KNN again slightly behind the others. Again, for XGB and LGBM, calibration has only a slight effect on calibration error. For KNN, SVM, and RF, calibration is more beneficial, although the calibration error for KNN and RF does somewhat increase with the shift. The estimation and calibration errors are even more strongly correlated, as shown in Table~\ref{correlation}, especially for the AC and DoC-Feat variants. 

\section{Discussion}\label{discussion}

In this section, we analyze the significance of our findings. We underline some properties of confidence-based estimators in general and discuss the limitations and validity of our results.

\subsection{Novel Contributions}

Earlier studies~\shortcite<e.g.,>{guillory:2021,garg:2022,deng:2023,baek:2022,chen:2021,lu:2023} have explored the problem of unsupervised accuracy estimation only on the dataset level, that is, whether the suggested methods would be able to correctly estimate the accuracy of an ML model on some given shifted dataset, in the absence of GT labels. Furthermore, theoretical guarantees for the proposed methods are provided only by~\shortciteA{chen:2021} and~\shortciteA{garg:2022}. In real-life model monitoring, instead of complete, shifted datasets, users typically have access only to relatively small samples of the potentially shifted data and are mostly interested in detecting relevant changes in the underlying data distribution on the fly to trigger appropriate interventions. In this study, we have tried to bridge this gap by exploring different confidence-based performance estimators in a monitoring setting with relatively small sample sizes.

A minimal requirement for any usable accuracy estimator is that the mean of absolute estimation error is lower than the actual change in accuracy. Even then, variance in the estimates and sample accuracies due to sampling effects make monitoring based on any confidence-based estimator tricky. However, if on average, the estimation error is small enough and the sample accuracies do not vary too much, it is possible to detect consistent changes in model performance. These properties are realized in many cases in our experiments, underlining that the compared estimators can provide a feasible approach to model monitoring at least in some scenarios. Another encouraging finding is that for many cases, the estimation errors remain rather stable even under severe shift (although their variance slightly increases), which means that the probability of triggering an alert increases when the shift in performance increases. However, there are cases where the estimation errors are too big to be of any practical use. A user needs to carefully assess the applicability of any performance estimator in a case-by-case manner.

One of the most interesting findings in this paper is the strong correlation between many of the estimators and the calibration error of the models being estimated. This correlation is most prominent with AC, especially with calibrated confidence scores. Earlier studies have not reported calibration errors for the confidence scores either before or after applying a calibration mapping. We hypothesize that this shortcoming might explain some of the contradicting results between said studies on the usefulness of applying calibration. We suggest that all future studies on confidence-based accuracy estimators should report the calibration error of their models to make the results more comparable and to provide more insight into the potential success of the suggested methods.

\subsection{Remarks on AC and Other Confidence-based Estimators}

Ever since AC was introduced as a method for OoD-detection~\shortcite{hendrycks:2017}, it has become a de facto baseline for confidence-based estimators. However, up until now, there has never been a proper justification for its use as such or a thorough analysis of its properties. As confidence-based estimators (such as CBPE in NannyML) are already being implemented in monitoring libraries intended for industrial use, such theoretical analysis is long overdue. In this paper, we have provided such a justification by showing that under the calibration assumption, AC is an unbiased and consistent estimator of model accuracy. Another desirable property of AC is that one can leverage the properties of the Poisson binomial distribution to derive CIs for the estimates, which are shown to be valid under the same assumption. None of the other methods have been reported to have this property. 

Contributions to the practical utility of AC by the theoretical and experimental findings presented in this paper are three-fold. First, they guarantee that if a model is known to be calibrated, a user can confidently rely on AC to produce good quality estimates of model accuracy on average. Second, the deviations in these estimates can be accounted for using CIs, which are also guaranteed to be valid under the calibration assumption. Third, the strong correlation between calibration and estimation errors for AC shows that using AC in unsupervised accuracy estimation reduces the problem of estimation into a problem of calibration, which is a widely studied and constantly developing research area.

Although using control limits and CIs to account for sampling effects and uncertainty in the estimates respectively, two other sources of uncertainty remain. First, in this work, we have treated the target labels used in model training and calibration as reliable representations of the GT. In real-life applications, this might not be the case, and measures to deal with label noise might be needed. Second, the confidence scores are only point estimates of the empirical probabilities. A more thorough treatment would account for the uncertainty in these estimates as well. We leave the details of this treatment for future work.

Although this study has shown that the estimates produced by AC can compete with those provided by more complex methods and in many cases even beat them, the results and the theoretical analysis of AC should not be taken to imply that it is some magical bullet for all unsupervised monitoring problems. As stated by~\shortciteA{garg:2022}, no estimator can prevail under concept shift and no estimator is the best in all cases. Our experiments provide an empirical demonstration of this. Without access to GT labels, there is no reliable way to know if concept shift is happening. Thus, confidence-based estimators should only be relied on in scenarios that can be expected to contain no concept shift.

Confidence-based estimators may also struggle in situations where the target distribution contains data outside the support of the source distribution since models can not be expected to produce accurate confidence scores for areas of the input space they have not previously encountered. Thus, in cases where previously unseen areas of the input space might be introduced during inference, we suggest that any confidence-based estimator should be complemented with input data monitoring to recognize such anomalies. 

\subsection{Limitations and Validity}

In our experiments, we relied solely on synthetic datasets. Earlier studies on unsupervised accuracy estimation have focused on image classification with a minor interest also in textual data. In this work, we extend the scope of analysis to tabular data. Unfortunately, to the best of our knowledge, there are no real-life tabular datasets for covariate shift scenarios. Only recently, a benchmark for exploring distributional shifts in tabular data was published~\shortcite{gardner:2024}, but all of the datasets in this benchmark express concept shift, rendering them unusable for our purposes.

Since the AC method hinges on the calibration assumption, it is worthwhile to ponder how reasonable this assumption is in reality. An often cited study~\shortcite{ovadia:2019} shows that the uncertainty estimates of most models deteriorate with covariate shift. However, the study in question deals only with multi-class image classification and equates calibration with Temperature Scaling~\shortcite{guo:2017}. It is possible that the same conclusion might not hold for other scenarios or calibration techniques. In fact, promising techniques on how to remain calibrated under covariate shift are already starting to emerge~\shortcite{białek2024}. Our experiments show that if no interventions are made, the calibration error seems to increase slightly but steadily with increasing shift, also in a binary classification setting using Isotonic Regression~\shortcite{zadrozny:2002}. We also experimented with Beta Calibration~\shortcite{kull:2017} to derive the calibration mappings with similar results. However, although correlated, the estimation errors do not seem to increase as rapidly as the calibration errors do, implying that the performance of the estimators remains reasonable even if the calibration error did increase with shift. 

Although the calibration errors for XGB and LGBM were almost unaltered by the increasing shift with both signaling low calibration errors even without applying a calibration mapping, this effect might be an artifact stemming from the simplicity of our synthetic datasets. In a more high-dimensional setting, gradient boosting can lead to high calibration error as well~\shortcite{kivimaki:2023}. However, some models or ways to train models might be more robust against shift-induced miscalibration. We speculate that this might also provide a more credible explanation for why in~\shortcite{guillory:2021}, the (regression-free) AC method performs better than DoC with Augmix-DeepAugment since training with Augmix is known to reduce miscalibration~\shortcite{hendrycks:2020}. Although perfect calibration is unattainable in real-life scenarios and calibration tends to deteriorate under shift, we encourage future research to explore further methods to help maintain sufficient calibration under covariate shift, enhancing the performance of confidence-based estimators. 

\section{Conclusion}\label{conclusion}

In this study, we revisited AC, a commonly used baseline method in unsupervised accuracy estimation. We analyzed its theoretical properties under the assumption of perfect calibration and showed it to be an unbiased and consistent estimator of the predictive performance of a deployed ML model. We showed experimentally that by leveraging the properties of the Poisson binomial distribution, it is possible to find valid confidence intervals for the estimates provided by AC, which is a property no other confidence-based estimator currently has to the best of our knowledge. We compared the quality of the estimates provided by AC against more advanced confidence-based estimators and experimentally showed that despite its simplicity, AC is competitive against these estimators, as long as the calibration error remains small enough. In addition to potential calibration error, encountering concept shift or data outside the support of training data might hamper the performance of AC when monitoring a model in production. Extending the scope of confidence-based performance estimation to include other metrics in addition to accuracy and comparing the estimators in real-life scenarios are important challenges, which we leave for future research.

\acks{This work was partly funded by local authorities (“Business Finland”) under grant agreement 20219 IML4E of the ITEA4 programme.}

\appendix
\section*{Appendix A. Data Description for the Covariate Shift Experiment}\label{a_data_description}

In this appendix, we describe the data used in the covariate shift experiment described in Section~\ref{covariate_shift}.  

\subsection*{A1. Data Used in the First Scenario}

In the first scenario, we drew samples from a mixture of six 2-dimensional Gaussian distributions. The labels were assigned stochastically according to their distance from the line $y=x$. Points on the line had a $50\%$ probability of being assigned with a label 1. For points below the line, this probability gradually increased and for points above the line, it decreased, as dictated by the function $p(\boldsymbol{x})=\frac{1}{1+e^{-\gamma d}}$, where $d$ is the signed distance of $\boldsymbol{x}$ from the line $y=x$, and $\gamma$ controls the rate for which the probability changes with $d$. In the reported experiments, we used $\gamma=\sqrt{2}$ to get a noticeable yet realistic decay in accuracy for different degrees of shift. 

\begin{figure*}[ht!]
\centering
\includegraphics[width=1.0\textwidth]{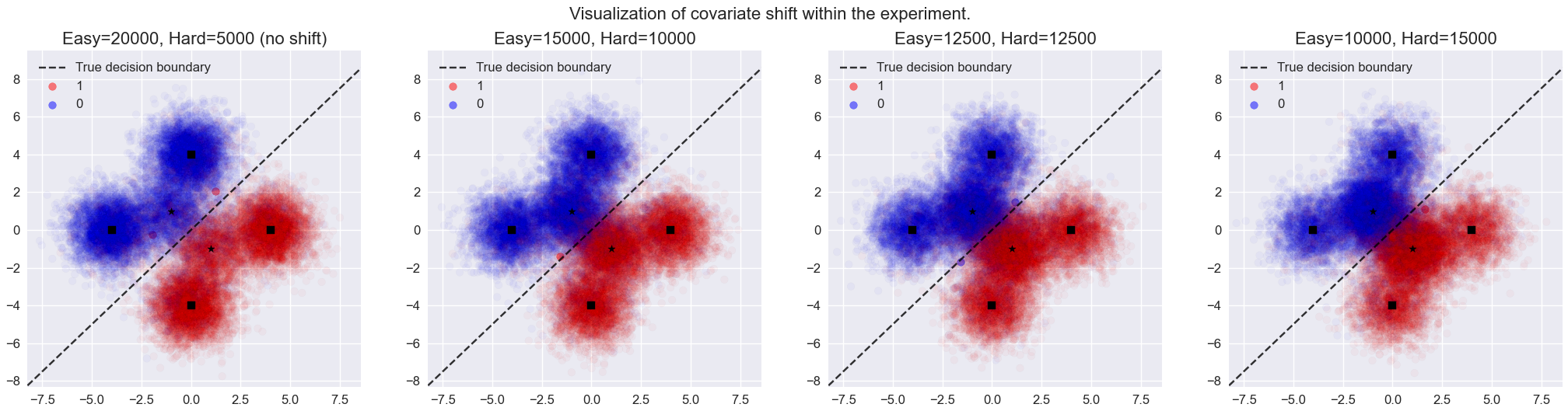}
\caption{A visualization of the covariate shift with linear decision boundary. The dashed line signs the decision boundary of the Bayes optimal classifier. The modes of the hard-to-predict mixture components are marked with '$\star$' and the modes of the easy-to-predict components are marked with '$\blacksquare$'.}
\label{data_gradient}
\end{figure*}

\begin{figure*}[ht!]
\centering
\includegraphics[width=1.0\textwidth]{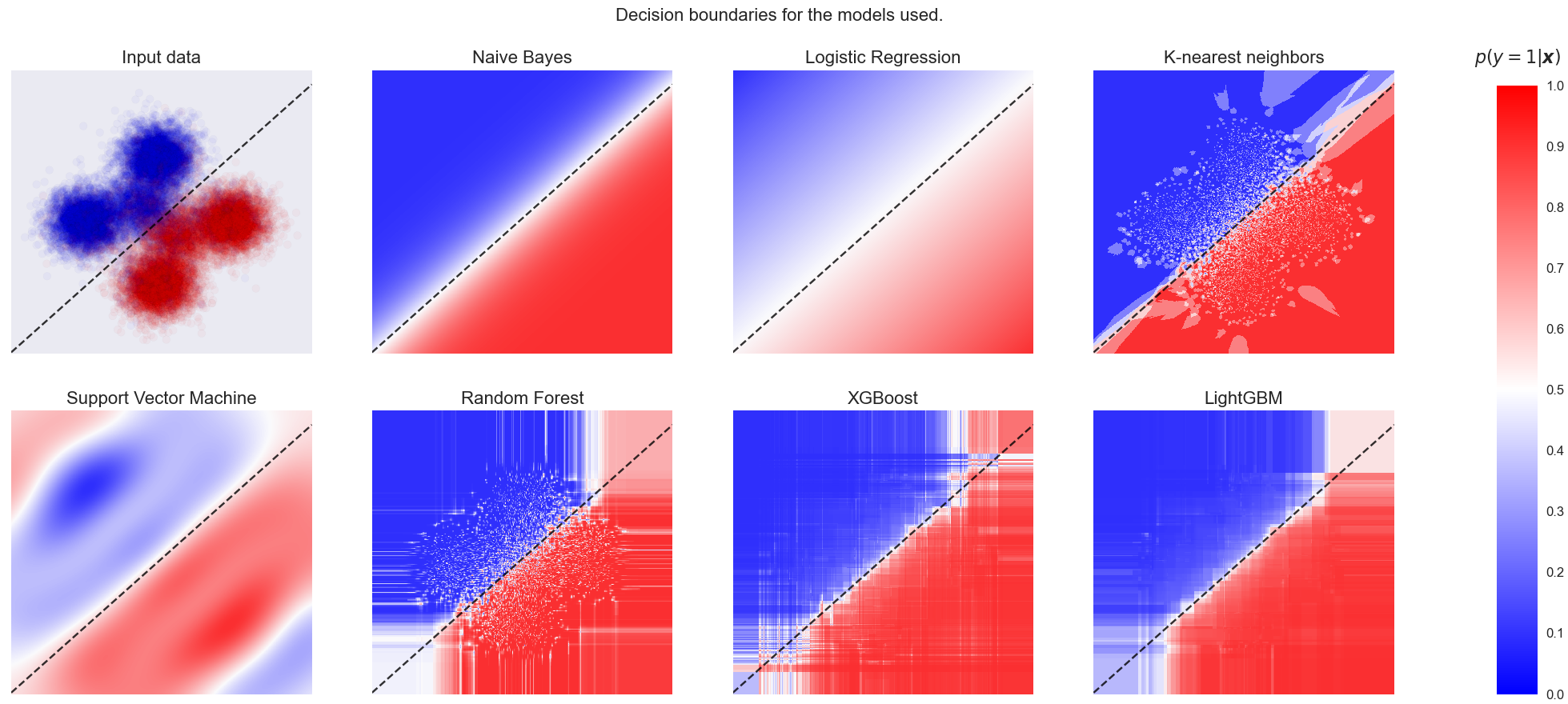}
\caption{The predicted distribution of $p(y=1|x)$ for all models in scenario 1. The decision boundary used by each classifier is where they assign $p(y=1|x)=0.5$ (the white region). The dashed reference line marks the decision boundary of the Bayes optimal classifier.}
\label{db_gradient}
\end{figure*}

The decision boundary for the Bayes optimal classifier was exactly the line $y=x$. The modes of the six mixture components were arranged symmetrically to the line $y=x$ so that two of them lay closer to the decision boundary on opposite sides at $(1,-1)$ and $(-1,1)$, and the other four a bit further with two on each side at $(4,0),(0,-4),(-4,0)$, and $(0,4)$. All components had equal diagonal covariance matrices $c\textbf{I}$, where we used $c=1$ to control the dispersion of data points sampled from each component. We sampled data equally from both sides of the line $y=x$ to get a balanced set of labels (up to sampling effects). A visualization of the datasets for each degree of shift is given in Fig.~\ref{data_gradient}. The resulting predicted probability distributions for each trained model are illustrated in Fig.~\ref{db_gradient}, showing their respective inductive biases. 

\subsection*{A2. Data Used in The Second Scenario}

In the second scenario, we drew samples from a mixture of nine 2-dimensional Gaussian distributions. Again, the labels were assigned stochastically, but this time according to their distance from the circle $x^2+y^2=5$. Points on the circle had a $100\%$ probability of being assigned with a label 1. This probability decreased gradually as the distance from this circle increased, according to the function $p(\boldsymbol{x})=e^{-\gamma d^2}$, where $d$ is the signed distance of $\boldsymbol{x}$ from the circle $x^2+y^2=5$. Again, $\gamma$ controls the rate for which the probability changes with $d$. In the reported experiments, we used $\gamma=\log\sqrt{2}$ to get a roughly similar decay in accuracy for both scenarios.

\begin{figure*}[ht!]
\centering
\includegraphics[width=1.0\textwidth]{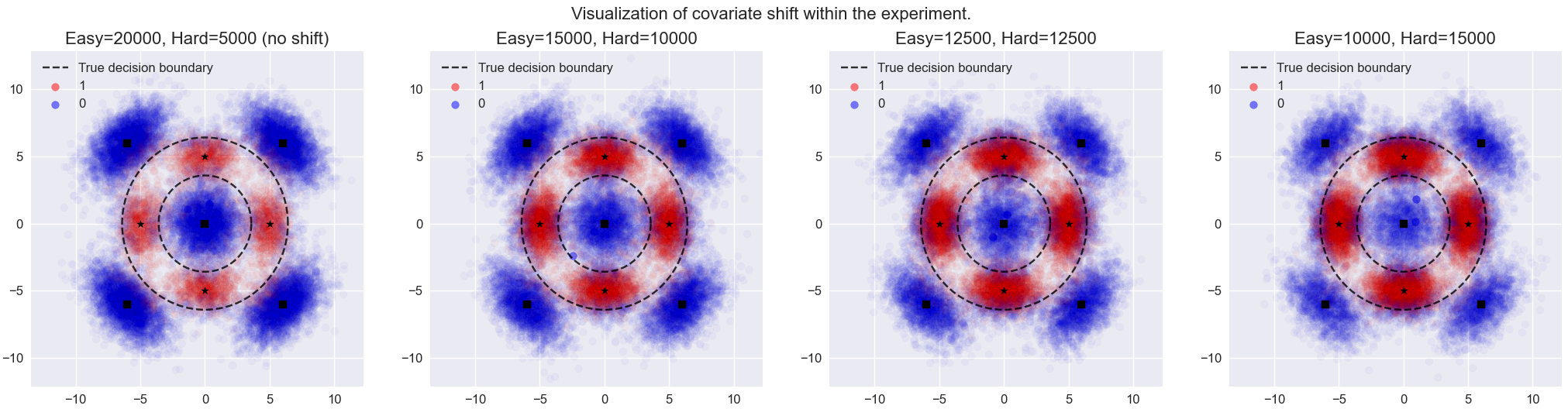}
\caption{A visualization of the covariate shift with non-linear decision boundary. The dashed line signs the decision boundary of the Bayes optimal classifier. The modes of the hard-to-predict mixture components are marked with '$\star$' and the modes of the easy-to-predict components are marked with '$\blacksquare$'.}
\label{data_circular}
\end{figure*}

\begin{figure*}[ht!]
\centering
\includegraphics[width=1.0\textwidth]{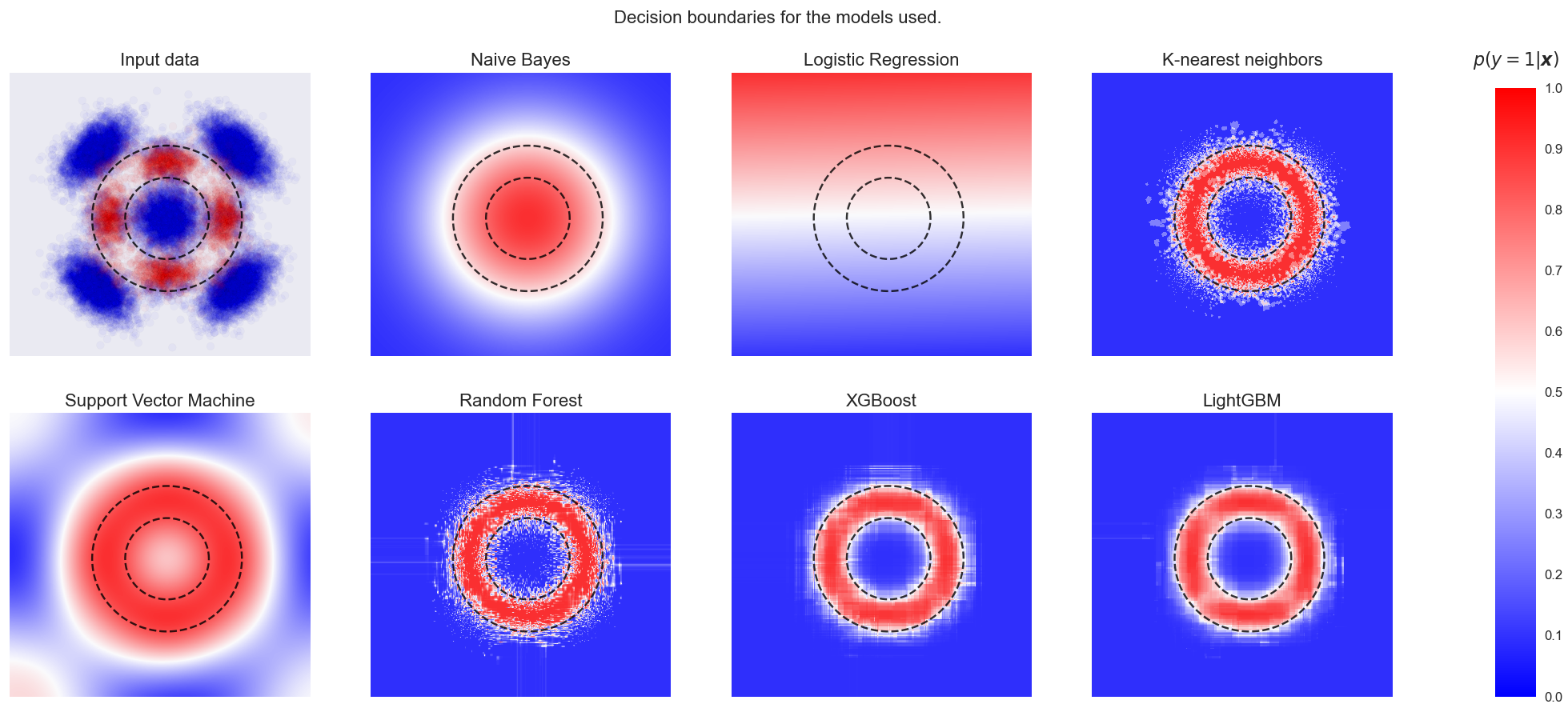}
\caption{The predicted distribution of $p(y=1|x)$ for all models in scenario 2. The decision boundary used by each classifier is where they assign $p(y=1|x)=0.5$ (the white region). The dashed lines mark the decision boundary of the Bayes optimal classifier.}
\label{db_circle}
\end{figure*}

As a result, the decision boundary for the Bayes optimal classifier consisted of two concentric circles, where the Bayes optimal classifier assigned label 1 to each point in between these two concentric circles and label 0 to each other point. Again, the modes of the mixture were arranged symmetrically, so that the models could learn a reasonable decision boundary. One of the easy modes was set at the origin $(0,0)$ and the other four outside the outer circle at $(6,6),(-6,-6),(-6,6)$, and $(6,-6)$. The four hard modes were placed in between the concentric circles at $(5,0),(-5,0),(0,5)$, and $(0,-5)$. To cover more area around the optimal decision boundaries, we replaced diagonal covariances for all but the component with its mode at the origin with more complex covariance matrices. For the other four easy components, we used covariance matrices
$\left[
    \begin{matrix}
        2&-1\\
        -1&2
    \end{matrix}
\right]$
for the first two, and
$\left[
    \begin{matrix}
        2&1\\
        1&2
    \end{matrix}
\right]$
for the other two components respectively. For the hard components, we used 
$\left[
    \begin{matrix}
        1&0\\
        0&2
    \end{matrix}
\right]$
for the first two, and
$\left[
    \begin{matrix}
        2&0\\
        0&1
    \end{matrix}
\right]$
for the other two respectively as the covariance matrices. A visualization of the datasets for each degree of shift is given in Fig.~\ref{data_circular}. The resulting predicted probability distributions for each trained model are illustrated in Fig.~\ref{db_circle}, showing their respective inductive biases. 

Contrary to the first scenario, this way of sampling induced a label shift, with an increasing fraction of positive labels as shown in Table~\ref{label_shift}.
\begin{table}[!ht]
    \centering
    \begin{tabular}{c|c} 
        Shift & \% of positives \\ \hline
        0 & 19.86 \\
        1 & 34.22 \\
        2 & 40.62 \\
        3 & 48.20 \\
        \hline
    \end{tabular}
    \caption{The fraction of positive labels in the second scenario.}
    \label{label_shift}
\end{table}
However, we did not consider this to be a source of error given that the model performance deterioration rate was very similar to what was observed in the case of a linear decision boundary.

\vskip 0.2in
\bibliography{JAIRAuthorKit/bibliography}
\bibliographystyle{theapa}

\end{document}